%% file: LefevrePIEEE2017.tex
\documentclass[draftcls,12pt,onecolumn]{IEEEtran} 

\usepackage{cite}

\ifCLASSINFOpdf
\usepackage[pdftex]{graphicx}
\graphicspath{{figures/}}
\DeclareGraphicsExtensions{.pdf,.jpeg,.jpg,.png}
\else
\usepackage[dvips]{graphicx}
\graphicspath{{figures/}}
\DeclareGraphicsExtensions{.eps}
\fi
\usepackage{hyperref}

\usepackage{amsmath}
\usepackage{array}

\ifCLASSOPTIONcompsoc
  \usepackage[caption=false,font=normalsize,labelfont=sf,textfont=sf]{subfig}
\else
  \usepackage[caption=false,font=footnotesize]{subfig}
\fi

\usepackage{url}
\usepackage{color}

\definecolor{MyDarkBlue}{rgb}{0,0.8,0.5} 
\definecolor{MyDarkOrange}{rgb}{0.8,0.5,0.0} 
\definecolor{pink}{RGB}{255,20,147} 
\definecolor{red}{RGB}{255,0,0}

\begin{document}

\newcommand{\mytitle}{

Towards seamless multi-view scene analysis from satellite to street-level
}
\title{\mytitle}

\author{S{\'e}bastien Lef{\`e}vre, \and Devis Tuia,~\IEEEmembership{Senior Member,~IEEE}, \and Jan D. Wegner, \and Timoth{\'e}e Produit, \and Ahmed Samy Nassar

\thanks{\noindent \textbf{This is the pre-acceptance version, to read the final, published version, please go to the DOI: \href{http://ieeexplore.ieee.org/document/7926323/}{10.1109/JPROC.2017.2684300}}.}
\thanks{\noindent SL is with IRISA institute, Universit\'e Bretagne Sud, Vannes, France. Email: sebastien.lefevre@irisa.fr, web: http://www.irisa.fr/obelix}
\thanks{\noindent DT was with the MultiModal Remote Sensing group, University of Zurich, Switzerland. He is now with the Laboratory of GeoInformation Science and Remote Sensing, Wageningen University \& Research, the Netherlands. Email: devis.tuia@wur.nl, web: http://www.geo-informatie.nl}
\thanks{\noindent JDW is with the Photogrammetry and Remote Sensing group, ETH Zurich, Switzerland. Email: jan.wegner@geod.baug.ethz.ch}
\thanks{\noindent TP is with G2C institute, University of Applied Sciences, Western Switzerland. Email: timothee.produit@heig-vd.ch}
\thanks{\noindent ASN is with IRISA institute, Universit\'e Bretagne Sud, Vannes, France. Email: ahmed-samy-mohamed.nassar@irisa.fr}
}

\markboth{Proceedings of the IEEE, DOI: 10.1109/JPROC.2017.2684300}{Lef{\`e}vre et al.: \mytitle; DOI: 10.1109/JPROC.2017.2684300}

\maketitle

\begin{abstract}
In this paper, we discuss and review how combined multi-view imagery from satellite to street-level can benefit scene analysis. Numerous works exist that merge information from remote sensing and images acquired from the ground for tasks like land cover mapping, object detection, or scene understanding. 
What makes the combination of overhead and street-level images challenging, is the strongly varying viewpoint, different scale, illumination, sensor modality and time of acquisition. Direct (dense) matching of images on a per-pixel basis is thus often impossible, and one has to resort to alternative strategies that will be discussed in this paper.
We review recent works that attempt to combine images taken from the ground and overhead views for purposes like scene registration, reconstruction, or classification. 
Three methods that represent the wide range of potential methods and applications (change detection, image orientation, and tree cataloging) are described in detail. 
We show that cross-fertilization between remote sensing, computer vision and machine learning is very valuable to make the best of geographic data available from Earth Observation sensors and ground imagery. Despite its challenges, we believe that integrating these complementary data sources will lead to major breakthroughs in Big GeoData.
\end{abstract}

\IEEEpeerreviewmaketitle

\section{Introduction}\label{sec:intro}
\input{intro}

\section{Literature survey}\label{sec:survey}

Ground-to-aerial matching of images or point clouds has received much attention in research, but is still largely unsolved. In the following, we roughly subdivide literature into three main application scenarios that call for different approaches to the problem: geolocalization, object detection and reconstruction, and multimodal data fusion. 

\subsection{Geolocalization}
\input{loc.tex}

\subsection{Matching and reconstruction}
\input{3D.tex}

\subsection{Multimodal data fusion}
\input{mm.tex}

\section{Selected examples}\label{sec:examples}
We describe three methods in detail to show how possible solutions for combining ground and aerial imagery may look like for specific tasks. Three systems that respectively aim to perform change detection, image geolocalization and orientation, and tree cataloging are explained and remaining challenges are highlighted.

\subsection{Multimodal change detection between ground-level panoramas and aerial imagery}\label{subsec:change}
\input{aertpd.tex}

\subsection{Terrestrial image orientation in landscape areas}
\input{PEP-ALP}

\subsection{Cataloging street trees with deep learning and Google Maps}\label{subsec:treestuff}
\input{registree}

\section{Conclusion}\label{sec:conclusion}
\input{conclusion}

\section*{Acknowledgment}

This work was partly funded by the Swiss National Science Foundation under the grant PP00P2-150593 and by the French Agence Nationale de la Recherche (ANR) under reference ANR-13-JS02-0005-01 (Asterix project).

\ifCLASSOPTIONcaptionsoff
  \newpage
\fi

\bibliographystyle{IEEEtran}
\bibliography{IEEEabrv,refs,DevisAddsOn}

\begin{IEEEbiography}[{\includegraphics[width=1in,height=1.25in,clip,keepaspectratio]{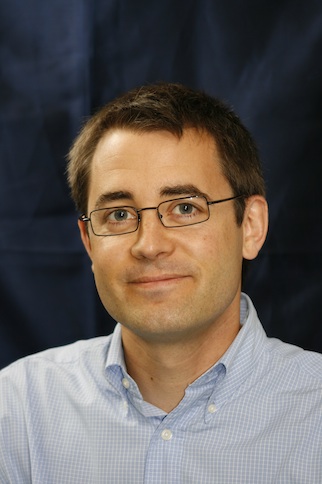}}]{S\'ebastien Lef\`evre}
received the M.Sc. in computer engineering from the University of Technology of Compi\`egne, France, in 1999, and the Ph.D. degree in computer science from the University of Tours, Tours, France, in 2002. In 2009, he received the French HDR degree in computer science from the University of Strasbourg, Strasbourg, France. 
From 2003 to 2010, he was an Associate Professor with the Department of Computer Sciences and the Image Sciences, Computer Sciences and Remote Sensing Laboratory (LSIIT), University of Strasbourg - CNRS. From 2009 to 2010, he was an INRIA invited scientist within the TEXMEX team of IRISA/INRIA Rennes. In 2010, he joined the Universit\'e Bretagne Sud, France, as a Full Professor in computer science, in the Institute of Technology of Vannes and the Institute for Research in Computer Science and Random Systems (IRISA), France. Within IRISA, he is leading the OBELIX team dedicated to image analysis and machine learning for remote sensing and earth observation. His research interests are in image analysis and pattern recognition, using mainly mathematical morphology, hierarchical models, and machine learning techniques with applications in earth observation and content-based image retrieval.
\end{IEEEbiography}

\begin{IEEEbiography}[{\includegraphics[width=1in,height=1.25in,clip,keepaspectratio]{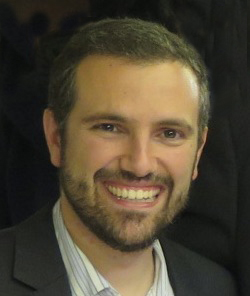}}]{Devis Tuia}
(S'07, M'09, SM'15)  received the Ph.D. from University of Lausanne in 2009.
He was a  Postdoc at the University of Val{\'e}ncia, the University of Colorado, Boulder, CO and EPFL Lausanne. From 2014 until 2017, he was Assistant Professor with the Department of Geography, University of Zurich. He is now associate professor with the Laboratory of GeoInformation and Remote Sensing, Wageningen University \& Research, the Netherlands. He is interested in algorithms for information extraction and data fusion of remote sensing images using machine learning. 
 More info on http://devis.tuia.googlepages.com/
\end{IEEEbiography}

\begin{IEEEbiography}[{\includegraphics[width=1in,height=1.25in,clip,keepaspectratio]{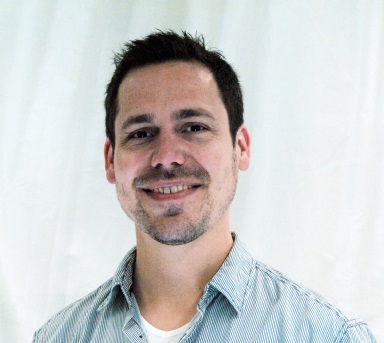}}]{Jan Dirk Wegner} received the Diploma degree (Dipl.-Ing.) in geodesy and geoinformatics and the Ph.D. (Dr.-Ing.) degree from Leibniz Universit\"{a}t Hannover, Germany, in 2007 and 2011, respectively. In December 2011, he was granted an ETH Postdoctoral fellowship and held a postdoctoral position with the Photogrammetry and Remote Sensing group, ETH Zurich, Switzerland,
between April 2012 and March 2015, where he has been a Senior Scientist since April 2015. He is mainly interested in geospatial computer vision, and large-scale machine learning, with applications to tree inventories, underwater 3-D reconstruction, and semantic 3-D reconstruction and segmentation. Together with colleagues, he is running the 1) ISPRS benchmark challenge for 2-D pixelwise semantic segmentation, object recognition, and 3-D reconstruction using very high resolution aerial images and airborne laser scans; and 2) the Large-Scale Point Cloud Classification Benchmark that contains terrestrial laser scans of urban scenes with over one billion labeled points. Dr. Wegner received the Science Award of the German Geodetic Commission in 2014, awarded every two years to an internationally recognized young researcher.
\end{IEEEbiography}

\begin{IEEEbiography}[{\includegraphics[width=1in,height=1.25in,clip,keepaspectratio]{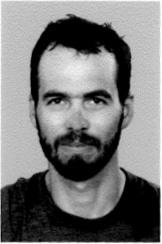}}]{Timoth\'ee Produit}
received a Master in environmental engineering  and the Ph.D. from EPFL, Lausanne, Switzerland in 2015. He is scientific collaborator at HEIG-VD, Yverdon-Les-Bains, Switzerland. He is interested in the interaction between pictures and virtual globes and geographical exploitation of public landscape images for environmental studies.
\end{IEEEbiography}

\begin{IEEEbiography}[{\includegraphics[width=1in,height=1.25in,clip,keepaspectratio]{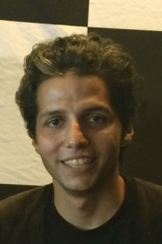}}]{Ahmed Samy Nassar}
received his B.Sc. degree in computer engineering from Nile University, Egypt. He is a master’s student and research assistant at the Center for Informatics Science at Nile University. His research interests focus is on computer vision, machine learning and image processing with applications in remote sensing. 
\end{IEEEbiography}

\end{document}

%% file: intro.tex
\IEEEPARstart{T}{raditionally}, interpretation of satellite and aerial imagery has been approached separately from scene analysis based on imagery collected at street-level. The reasons are partially historical, the Earth Observation and remote sensing community has worked on overhead images whereas the computer vision community has largely focused on mapping from data acquired at ground-level. In this paper we advocate to view these efforts as complementary. We will gain new insights if viewing them as one joint attempt to model our environment from multiple different viewpoints, at different spatial and temporal resolutions, and with different sensor modalities. We believe that all collected data, regardless of viewpoint and sensor modality, should fruitfully cooperate to create one, holistic scene representation. 

\paragraph*{Earth Observation}

Earth Observation has experienced a significant evolution during the last decades, due to advances in, both, hardware and software. Manual interpretation of analogue, aerial surveys that were limited to campaigns at city-scale, was the dominant technology for most of the XXth century \cite{Bel53}. The launch of satellite sensors for Earth Observation (EO) changed this situation and allowed a significant improvement in terms of spatial scale and temporal coverage. Since Landsat~1 (1972) and its 80-meter spatial resolution and 18-day revisit period, the ground sampling distance, image quality and availability of EO data has been growing rapidly. Today, state-of-the-art satellites offer decimetric accuracy (e.g., WorldView-2 and 3, Pleiades); their constellations offer weekly or even daily revisit cycles (e.g., RapidEye, Copernicus Sentinels). New swarm-like satellite schemes, that rely on a multitude of small, low-cost sensors are currently changing the field and will soon have an impact on how we work in EO (e.g., Google Terra Bella or PlanetLabs constellations).
This ever-increasing data flood can only be analyzed if relying on automated methods that allow close to on-the-fly image interpretation for such big data. Current scientific efforts in computer vision, pattern recognition, machine learning, and remote sensing are making considerable progress, especially since the comeback of deep learning.

\paragraph*{Street-level images and Social Media}

The massive amounts of geo-tagged images on social media (e.g., Flickr, facebook, Instagram) provide another possibility for big data solutions \cite{Luo11}. 
Flickr gathers 122 millions of users, sharing daily 1 million of photos, for a total of 10 billion shared images. Beyond generic photo-sharing platforms like Flickr, some geolocation-oriented solutions have been introduced in the last years. 
360cities provides a collection of high-resolution georeferenced panoramic photos.
Panoramio was a geolocation-oriented photo-sharing platform, offering 94 millions of photos (with 75 billions of views). It is now part of Google Maps, illustrating the convergence between ground-level imagery and remote sensing technologies. Mapillary aims to share geo-tagged photos in a crowdsourced initiative to offer an immersive street-level view of the world. Its 100 millions of photos are exploited with computer vision techniques in order to detect and recognize objects, perform 3D modeling or text/sign understanding.

In addition to crowd-sourced geo-tagged images, Google Street View \cite{Vin07}, Bing Maps Streetside, or Apple Maps are offering ground-level views that are acquired with planned mobile mapping campaigns at worldwide scale. Dedicated imaging campaigns guarantee much denser and complete data coverage (i.e., all street scenes of a city) as opposed to most crowd-sourcing efforts that usually do not have much success beyond tourist sights. Imagery acquired with large campaigns usually facilitates processing because data is provided in a homogeneous format and exhibits less illumination change. In addition to worldwide imaging campaigns, there are many similar initiatives at the national or regional scale, e.g. Tencent Maps in China, Daum Maps in South Korea, Mappy in France, GlobalVision VideoStreetView in Switzerland, CycloMedia in various European countries and US cities, etc.

Geo-tagged ground-based images shared via social media (captured by pedestrians, cars, etc.) thus appear as a relevant complementary data source for EO applications. These data provide a high-resolution viewpoint from the ground that offers much potential to add missing information to overhead imagery to map and monitor natural and human activities on Earth. Furthermore, it allows circumnavigating typical restrictions of remotely sensed data: it is not affected by cloud coverage nor limited to nadir or oblique views, and it is widely available via the pervasive use of mobile phones, which allows for immediate feedback. 
Nonetheless, this imaging modality comes with its own specific challenges, including a large variability in terms of acquisition conditions (leading to various perspective effects, illumination changes), frequent object occlusions or motions. Successful coupling of ground-level and remote sensing imagery requires addressing their differences in viewpoint, sensor modality, spatial and temporal resolutions, etc.

\paragraph*{Contributions}

We present a thorough and detailed review of works that combine overhead and street-level images. Three projects are described in detail to complete this survey; each of these presents a different method for a geodata task, and emphasize the impact of cross-fertilizing computer vision and remote sensing. 
This paper is meant to survey existing methods for aerial-to-ground image combination, to highlight current challenges but also to point in promising directions for future work. It shall serve as a motivation for colleagues working in Earth Observation, remote sensing, machine learning and computer vision to more closely cooperate, to share ideas, and to show that there are many common interests. 

\paragraph*{Paper Organization}
The paper is organized as follows. We provide in Sec.~\ref{sec:survey} a survey of the literature in ground-aerial/spatial matching, and we review existing works considering three specific problems that are geolocalization, matching and reconstruction, and multimodal data fusion. These various problems are illustrated in Sec.~\ref{sec:examples} through an in-depth presentation of some selected examples, that correspond to some of the authors' previous works. The three presented systems aim at coupling ground imagery and remote sensing to perform change detection, image geolocalization and orientation, and tree cataloging respectively. Finally, we conclude this paper with Sec.~\ref{sec:conclusion}.

%% file: loc.tex
A great number of pictures found on photo-sharing platforms does not provide information about their spatial location. If available it is either derived from GPS of the device, or an approximate location is entered manually by the photographer at the time of upload. 

However, this large amount of images without geolocalization is available on social media and could be used, for example, to improve indexing, facilitate search and offer augmented reality experiences. For these reasons, locating pictures has become a fast moving field in computer vision, which has attracted attention in recent research. 

Early works focused on the use of the geolocated images to extract semantic geographic information and enrich global maps. Proximate sensing has been used by Leung and Newsam~\cite{Leu10} to build a land cover map of a $100 \times 100$ km region of England (with a 1 km spatial resolution). To do so, input pictures are extracted from the Geograph database based on their grid location. Such a system remains tributary of having all pictures georeferenced. It is also possible to match photos that do not have geotag with those in a database made of geolocated ones. The \emph{IM2GPS} system from Hays and Efros~\cite{Hay08} retrieves the pictures being the most similar to those in the database and is used to study whether these images are also geographically close in space or if they represent more spatially diffuse (and therefore difficult to precisely locate) types of land cover. Existing land cover and light pollution global maps (and their scores at the geolocated pictures' location) are also used to estimate land cover and light pollution in all the pictures.

However, the applicability of these approaches is reduced to popular locations (often tourist sights), where most web-shared images are taken and for which data availability (and even redundancy) can be guaranteed. One possibility to adapt geo-localization models to wider areas is matching ground-level photos to overhead images. Aerial or satellite images at very high-resolution are available almost everywhere on the planet and provide a much more complete and homogeneous image database. A true challenge that receives much attention in recent research is finding correspondences between overhead and ground-level views. Dense, pixel-wise image matching is literally impossible due to the great change in viewpoint. Overhead images and photo collections on the ground also differ greatly in resolution, illumination etc., which calls for very robust, truly multimodal approaches to correspondence search across aerial and ground views. 

Recent works aim at developing feature extraction methods that define common representation spaces, in which a ground photo is projected close to the corresponding aerial image patch. Kernel canonical correlation analysis (KCCA~\cite{Lai00}) has been used by Lin et al.~\cite{Lin13} to learn the relationships between pictures and aerial images or land cover maps available online. The KCCA score is used to map geographically the similarity of a picture with appearance observed from above. This way, one can evaluate the confidence of the localization in the geographical space and for example discriminate pictures of vague environments (e.g. a beach) from pictures with strong location uniqueness (e.g. the Eiffel tower).
Deep neural networks play a key role in works aiming at image localization using overhead imagery (e.g.,~\cite{Wor15b}), with Siamese Networks being the most popular variant in current systems~\cite{Had06,Lin2015,Vo16}. We describe a method that relies on siamese CNNs as a workhorse to perform change detection in Sec.~\ref{subsec:change}.

In robotics, localization of a rover has been tackled by the joint use of overhead imagery and ground sensors embedded onto the vehicle. Among existing systems, we can cite the use of a ground LiDAR to detect tree stems in forest environment~\cite{Hus13}, where the detected candidates are then  matched with tree centers extracted from RGB aerial image analysis to locate the rover. Videos on the rover are compared to very high resolution grayscale images in \cite{Leu08}, and location is retrieved using a particle filter. Finally, Google Maps images can be matched with panoramic images from multiview cameras installed on the rover \cite{Vis14}, by creating a synthetic overhead view using the images from the ground cameras and performing keypoint matching.

%% file: 3D.tex

While geolocalization often requires only sparse correspondences across views, object detection and particularly reconstruction calls for more precise and often dense matching between data of different viewpoints. Per-pixel matching from aerial to ground is an extreme wide-baseline matching problem. Direct (dense) matching of aerial nadir-images and street-level panoramas is very hard, because it requires compensating for $90^\circ$ viewpoint changes, and usually also scale differences. Street-level images often have much higher resolution and show more details than aerial or satellite images. Much of the scene can only be observed from either street-level or overhead perspective, for example, building roofs can hardly be captured from street-level, but dominate building appearance in overhead images (the inverse counts for building facades). As a consequence, most approaches try to avoid per-pixel matching and propose late fusion of detection or reconstruction outcomes. Objects are detected or reconstructed separately per viewpoint, and results are combined.  

Nevertheless, many works aim at fine-grained 3D reconstruction of cities, where missing parts of the scene, invisible from the air, are added from street-level imagery. The ultimate goal is a complete, detailed, watertight 3D model (in mesh or voxel representation) at city-scale that allows seamless navigation in 3D.

To the best of our knowledge, the first attempt to this aim is the work of Fr\"{u}h and Zakhor~\cite{fruehcvpr2003,fruehcga2003}. They densely model building facades using a mobile mapping system equipped with laser scanners and a camera. Airborne laser scans are used to fill in missing data, like roofs and terrain. A digital surface model (DSM) is created from airborne scans, triangulated and textured with aerial images. Building facades modeled from street-level data are globally registered to the DSM derived from airborne scans using Monte-Carlo-Localization (a particle filter variant), which refines initial poses of street-level scans with respect to an edge map derived from the DSM.

An alternative method is proposed by Fiocco et al.~\cite{fiocco2005}, who combine terrestrial and airborne laser scans for the purpose of 3D city reconstruction. 
After manual coarse alignment of terrestrial and aerial scans, roof-edges are detected in both data sets, projected to the horizontal plane and matched to refine rotation and translation parameters. 
The combined point cloud is processed to a global triangle mesh, which is further filtered via a volumetric octree. The final mesh is extracted by contouring the iso-surface with a feature preserving dual contouring algorithm.

In contrast to the previous two approaches, Kaminsky et al.~\cite{kaminsky2009} do not combine multiple laser scans in 3D, but aim at aligning 3D terrestrial point clouds computed from images via structure-from-motion to a 2D overhead image. They estimate the optimal alignment with an objective function that matches 3D points to object edges in the overhead image. 
B\'odis-Szomor\'u et al.~\cite{bodisicpr2016} propose an approach that efficiently combines a detailed multi-view stereo point cloud with a coarser, but more complete point cloud acquired with an airborne platform to reconstruct one, joint surface mesh. At its core, the method does point cloud blending, by preferring points stemming from street-level data over airborne points, and volumetric fusion based on ray casting across a 3D tetrahedralization of the model. Bansal and Daniilidis~\cite{Ban14} match street view pictures to a 3D model of the skyline rendered by modeling the buildings using LiDAR scans, whereas Hammoud et al.~\cite{Ham13} use a set of modality-specific scores to match human annotations on the pictures with quantities extracted from a series of additional modalities: presence of points of interest from Open Street Maps, building cubes generated using LiDAR scans, land cover maps and estimation of the 3D skyline.

Other works rely only on optical images available worldwide (typically oblique and nadir images), for the pixel matching. Shan et al.~\cite{shan2014} propose a fully automated georegistration pipeline that can cope with ground to aerial viewpoint variation via object-specific, viewpoint-adaptive matching of key points. The approach mainly relies on planar, vertical building facades that can be observed in oblique aerial images, and street-level photos. 
Once ground images have been warped to oblique geometry, standard SIFT is applied to match across views. Similar approaches are found in~\cite{Ban11}, where building facades are detected in aerial oblique images and populate a database that is used to match facades detected in pictures; and in~\cite{Ban12}, where Bing's aerial views are matched with pictures from several sources (including Panoramio, Flickr and Google Street View) using a self-similarity index, where facades detected in the pictures are assigned to the closest cluster of facades extracted from the aerial view. Finally, Cham et al.~\cite{Cha10} locate Google Street view images using a vector layer of buildings footprints by finding corners and building outlines in both sources.

In mountain environments, the skyline has been used as a strong feature to locate pictures in virtual 3D environments rendered from Digital Elevation Models (DEM)~\cite{Bab11,Baa12}. The pictures, once located in the virtual 3D environment, can be used for digital tourism or be augmented with peaks' names and vector information. Such augmented panoramas can then be used for studying attractiveness for photography of the landscape, e.g. Chippendale et al.~\cite{Chi08} build a synthetic panorama from the global NASA digital terrain model around the area imaged in the picture and then align it to the pictures by matching salient points. Once the pictures have been located, an attractivness index is computed for every voxel in the synthetic image. Such index accounts for socially-derived aesthetics metrics, for example the number of views and comments a picture has received on the social media platform it was hosted on. An example of picture-to-landscape model matching in the Swiss Alps is presented in Section~\ref{sec:match}.

%% file: mm.tex
The previous two sections reviewed localization and matching by joint use of overhead and ground imagery. This section reviews recent research under the common name of \emph{multimodal data fusion}, a term often used in remote sensing for strategies using multiple different overhead sensors~\cite{Gom14} to map a certain scene. Here, this definition is extended to ground imagery. It includes approaches using ground pictures to enrich land cover classification or to improve object detection.

Land cover classification is probably the most studied area of remote sensing. Pictures from photosharing platforms might be used to validate the results of a land cover classification algorithm trained on overhead images~\cite{Fon15}: as an example, Foody and Boyd~\cite{Foo13} compare pictures centered at prediction locations to assess how accurate the GlobCover land cover layer is.
Other efforts were also reported for the classification of pictures retrieved from photo-sharing platforms as Flickr or databases as Geograph to describe land cover: proximate sensing~\cite{Leu10} was used to produce a map of the fraction of land developed in a $100 \times 100$ km region of England. The prediction is performed at a $1$ km resolution, by averaging  binary classification scores (developed vs. undeveloped) obtained for ground images that fall into a common grid cell of the Geograph database. The classifier used is an SVM trained on histograms of edge descriptors. In a follow up work \cite{Leu15}, the authors compared results obtained on the Flickr and Geograph pictures: they conclude that the Geograph results are more accurate, probably since the database is explicitly built to be geographically representative of land types in England. Works in~\cite{Zha12,Wan13} classify Flickr pictures to predict the presence or absence of snow and compare the pattern retrieved to those obtained by satellite-based products. The patterns look similar, but the poor spatial coverage of Flickr leads to a very sparse map compared to the one that can be obtained by satellite images. Estima and colleagues~\cite{Est13b,Est14} study the effects of such poor spatial coverage at the country level and conclude that these pictures are not distributed evenly enough to be used alone in a land cover classification effort thus calling for multimodal mapping systems.

Building up to that observation, some studies make successful use of both modalities (ground pictures and overhead images) to solve geospatial classification and object detection problem: Wegner et al.~\cite{wegner2016} apply a strategy for detecting individual trees at city-scale from street-level and aerial images. Tree detection is performed separately for all available street-level and overhead images, and all detections are projected to geographic coordinates followed by soft, probabilistic fusion (more details in Section~\ref{subsec:treestuff}).
Matthyus et al.~\cite{Mat16} propose a joint system to detect road observed in both street view and aerial images: using a structured SVM~\cite{Bak07} they learn a set of potentials accounting for 1) road smoothness, 2) lanes size, 3) the fact that roads are often parallel to each other, 4) they enforce consistency between detections in both modalities and 5) with Open Street Maps centerlines. Their results on the KITTI dataset~\cite{Gei13} augmented with aerial images provide impressive results against  state-of-the-art detectors.

Going beyond the classical tasks of land cover classification and object detection, an increasing interest is being observed on tasks related to the prediction of city attributes~\cite{Doe12,Zho14}. Pictures acquired on the ground can be used to predict activities (recreational, sports, green spaces), feelings of security, crime rates and so on. Such attributes are generally learned from the pictures themselves via visual attributes~\cite{Doe12} and existing image databases, which provide tags related to the activities~\cite{Zho14}. The attributes are then mapped in space, but few works make direct use of overhead images to provide complete maps carrying information where no pictures are available (we exclude the idea of direct spatial interpolation used in~\cite{Ari14}):  in~\cite{Wor15}, pretrained CNN models learned on picture databases (ImageNet and Places) are then applied on satellite image patches in order to see if similar geographical regions of the images were being activated by the same filters; in \cite{Lee15}, authors predict a set of city attributes (digital elevation, land use, but also population density, GDP and proportion of infant  mortality) by training classifiers on Flickr pictures labeled by the scores found at their geolocation on GIS or remote sensing maps. In \cite{Pro14}, authors use a One-Class SVM~\cite{Sch99} to predict the suitability of a spatial location for beautiful pictures taking: the model relates a set of geographical features (extracted from a DEM or open GIS layers) to the density of pictures found on Panoramio at that location. In \cite{Luo08}, authors classify activities happening in the pictures using two models based on handcrafted features  and fusing the classification scores at the end.\newline
A final set of works aims at transferring attributes from the pictures domain to the overhead images domain, in a domain adaptation~\cite{Tui15b} setting. Attributes from the pictures' domain would allow to obtain labels for activities unseen (and unrecoverable) in the images, as for example the type of activity being pursued on a grassland. Studies~\cite{Sun13,Sun15} search for common embedding spaces (for instance using subspace alignment) where images are mapped close to pictures depicting the same visual features. Once the mapping is done, the attributes of the nearest neighbours can then be transferred.

%% file: aertpd.tex
Despite the proliferation of Earth Observation programs, the endeavor of updating aerial or satellite high-resolution imagery is found to be quite costly and time-consuming for geographical landscapes around the world particularly when they are constantly and rapidly changing. Consequently, this acts as a limit towards maintaining an updated data source. We provide an efficient and low-cost solution by relying on georeferenced panoramic photos from digital cameras, smartphones, or web repositories as they offer high spatial information of the location queried. 

With Volunteered Geographic Information (VGI), humans are able to be extremely beneficial to geographical information systems by acting as intelligent sensors with a smartphone equipped with a GPS and a camera. Georeferenced photo collections are opening doors to ``proximate sensing'' as recalled in previous section. Many web services support georeferenced information: blogs, wikis, social network portals (e.g. Facebook or Twitter), but also community contributed photo collections discussed in the introduction of this paper. Such collections also benefit from the recent rise of affordable and consumer-oriented Virtual Reality (VR) headsets, in addition to 360$^{\circ}$ camera mounts and rigs that aim to generate panoramic spheres. As such, it will lead to the availability of an extensive collection of georeferenced panoramic photos since social networks as Facebook started lately adopting the format, and many uploaders' goal is to create the experience for the VR user as being in a certain location, which makes geolocating the panoramic photos straightforward.

In this section, we describe a recent multi-modal approach \cite{ghouaiel2016} to change detection, that makes use of both ground-level panoramic photos and aerial imagery. Furthermore, we introduce here several improvements to the existing method, that exploit deep neural networks to improve the overall results.

\subsubsection{Method}

\begin{figure}[ht]
\centering
\includegraphics[width=0.8\linewidth]{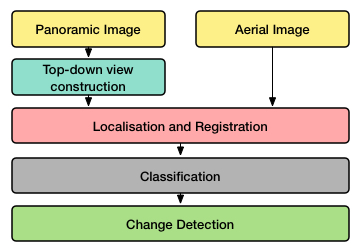}
\caption{Flowchart of proposed method.
 \label{fig:aertpdmeth}}
\end{figure}

The proposed method is summarized in Fig.~\ref{fig:aertpdmeth}. We first examine how to transform the georeferenced ground-level 360$^{\circ}$ panoramic photos (panoramic spheres) to top-down view images to resemble being obtained from a bird's eye view. The transformed or warped photo acquired is then compared with its counterpart aerial image from the same geolocation. This comparison indicates whether a change occurred or not at that location. This prevents routinely updating large landscapes, and allows requesting such an update of the aerial image only when necessary.

\paragraph{Data acquisition}
Depending on randomly uploaded panoramic spheres on web repositories, or by human effort was not a viable option to build our dataset due to the requirement of a large number of panoramic spheres in order to train a neural network for a particular area. Accordingly, we used Google Street View as it is the most reliable and vastest georeferenced source available for panoramic spheres. However, a procedure had to be devised to be able to acquire the photos efficiently. Having the geocoordinates of the Google Street View panoramas, we were able to amass aerial imagery associated with the same coordinates from Bing Maps which was the aerial imagery data source.

The data acquisition process thus relies on 3 successive steps: i) using Open Street Maps (OSM), a region of interest is selected and downloaded through OSM exporting tool. OSM file is then parsed for geocoordinates of the routes, since Google Street View panoramas are most likely to be located on routes to save processing time looking elsewhere; ii) through Google Maps API and Bing Maps API, requests were sent to receive the IDs of the images from both services. Due to proximity and availability issues, many IDs were duplicated so map and reduce operations were performed, to eliminate redundancy and ensure having a pair for every coordinate; iii) finally, each pair of images was downloaded and manually verified. This last step is necessary to ensure that each image pertains to the other (i.e. correspond to the same location) and to label any observed change.

\paragraph{Top-down view construction}
The panoramic photos being used here, obtained from Google Street View or from other sources, go through a warping procedure to get a bird's eye view image as proposed in~\cite{xiao2012recognizing}. 
The first step is constructing a spherical 3D model of the panoramic image. World coordinates are then computed using an inverse perspective mapping technique. Finally, bilinear interpolation is used on the panorama pixels to be able to get the color for the ground location. 

\begin{figure}[ht]
\includegraphics[width=\linewidth]{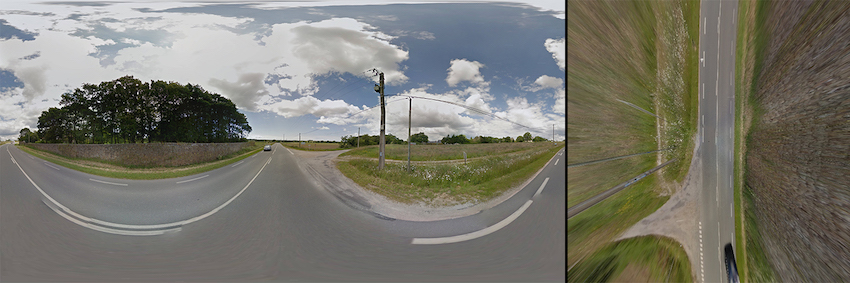}
\caption{Top-down view image (right) constructed from panoramic image (left).
 \label{fig:tpdconv}}
\end{figure}

An example of a top-down view image reconstructed from a panoramic photo is given in Fig.~\ref{fig:tpdconv}. While this process seems to be straightforward, its quality strongly depends on the kind of landscape considered. Indeed, top-down view construction for urban areas is quite problematic compared to suburban or rural areas due to how complicated the scene is. This causes many objects such as signs, or cars to blemish the output image, as well as objects towards the far end of the image get distorted heavily. Some illustrations are provided in Fig.~\ref{fig:sequenceofimages} (columns 1 and 2). 

\paragraph{Registration}
The top-down view image built in the previous step represents only a small portion of an aerial image to be compared with (e.g. Bing Map images are $150\times150 m^2$). Hence to compare correctly the images and detect changes, it is necessary to localize the top-down view image into the aerial image, then crop the area that contains the objects in the field of view of the Google Street View image. Similarly as already done in document image analysis~\cite{augereau2013semi}, we have explored  performance comparison of SIFT, SURF, FREAK, and PHOW for matching ground images to satellite images. The comparison has proven that SIFT is a superior method for the matching process, even when the satellite image contains many elements or is complicated. 

\begin{figure}[ht]
\includegraphics[width=\linewidth]{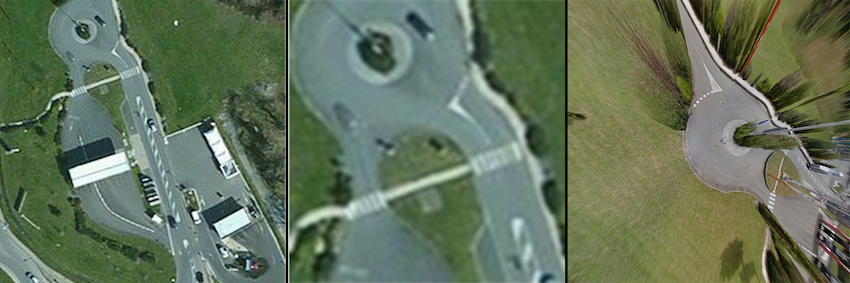}
\caption{Aerial localized image (middle) using top-down view image (right) on original aerial image (left).
 \label{fig:aertpd_3imgs_2c}}
\end{figure}

At first, SIFT~\cite{lowe2004distinctive} key points are extracted from both aerial and top-down images, and relevant descriptors are generated. Pairs of similar descriptors are found using Euclidean distance. The best match is then selected from the group of matches through a k-NN (nearest neighbor) classifier. We rely here on the FLANN library known for its good performance with k-NN search~\cite{muja2009fast}. Furthermore, to obtain geometric transformation between the matched key points, we use a homography matrix~\cite{agarwal2005survey}. To achieve this, we eliminate the outliers by RANSAC algorithm~\cite{fischler1981random}. Finally, if the required amount of inliers is achieved, which is a minimum of 4, the homography matrix could be computed.
An example of registration is provided in Fig.~\ref{fig:aertpd_3imgs_2c}.

\paragraph{Classification}

In our previous experiments~\cite{ghouaiel2016}, we have observed that the correlation coefficient called z-factor (see next step and \cite{ghouaiel2016}) was highly sensitive to the kind of landscape to be processed (i.e. urban and rural/suburban environments), making difficult the choice of an appropriate change detection threshold. 

Therefore, we introduce here an additional preclassification step to determine if the scene is rather urban or rural/suburban. To achieve this image classification task, we rely on a standard fine-tuned AlexNet deep architecture~\cite{krizhevsky2012imagenet} trained with Caffe~\cite{jia2014caffe} using a dataset of aerial images with an output of 2 classes (urban and rural). There was no need to classify the top-down view images as well, since the pair of images in the dataset are supposed to be of the same location, therefore of the same type.


\paragraph{Comparison}

Detecting the change between multiple images of the same location has been explored in previous works~\cite{shirvany2010maximum,xu2013change}. Our initial work~\cite{ghouaiel2016} was relying on a correlation index computed between the top-down view image and the related localized portion of the aerial image. A low correlation value indicates a change. To eliminate the cases in which some areas resulted in a very low correlation value, the z factor was calculated as proposed by~\cite{liu2011urban}.

Seeking to attain a better comparison detection accuracy, we explore here the use of Siamese Networks~\cite{chopra2005learning}. Such networks are most commonly used for face verification applications and have shown to be more accurate than hand-crafted features~\cite{lin2015learning}, with the goal of minimizing the feature vectors distance between matching images, and maximizing the distance for unmatching images.
The Siamese Network architecture is made of two identical Convolutional Neural Networks (CNN), each network being an AlexNet~\cite{krizhevsky2012imagenet} trained using Caffe~\cite{jia2014caffe}. During training, pairs of images (aerial or top-down view) are used to feed the two CNNs. The loss function layer used was the Contrastive Loss Function~\cite{hadsell2006dimensionality} to pull together matching pairs, and push far away unmatching pairs. The network architecture does not share weights or parameters considering each image is domain-specific~\cite{lin2015learning}. Each CNN generates a low dimensional feature representation or vector for the aerial and top-down view images, and Euclidean Distance is used to determine whether the views are close to each other or not. Changes are detected by setting a margin or threshold.

\subsubsection{Results and discussion}

\begin{figure*}[ht]
\centering
\includegraphics[width=1.0\textwidth]{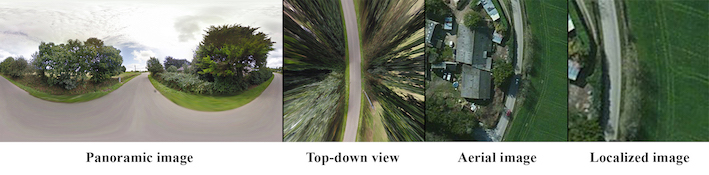}
\includegraphics[width=1.0\textwidth]{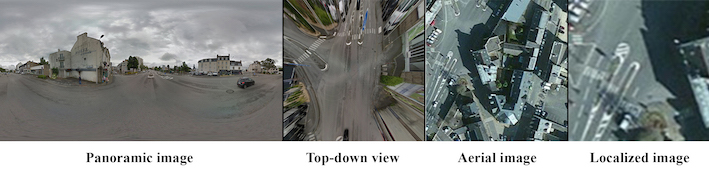}
\includegraphics[width=1.0\textwidth]{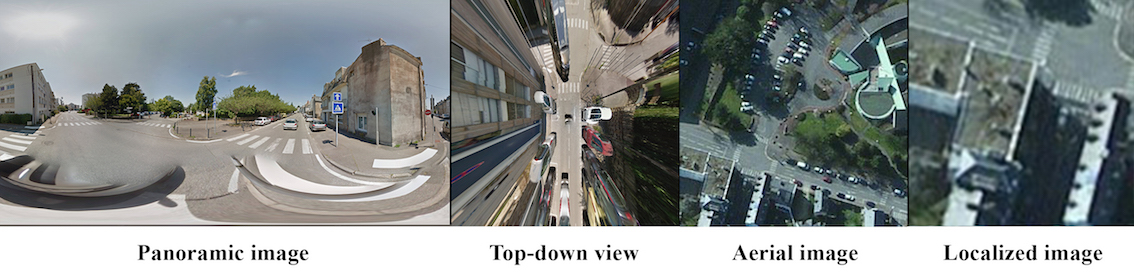}
\caption{Each collection above shows a panoramic image and its construction into a top-down view image, followed by an aerial image, and the localized part in the aerial image. While the top (rural) and middle (urban) collections are correctly addressed by the proposed approach, the bottom collection illustrates a challenging situation occurring when urban scenes contain many visible objects in the panoramic image.}
\label{fig:sequenceofimages}
\end{figure*}

The testing dataset was composed of 8,000 image pairs of aerial images acquired from Bing Maps and corresponding top-down views built from Google Street View panoramic images. The dataset spans four different cities (Vannes, Rennes, Lorient, Nantes) of Brittany, France. It contains both rural and urban areas. All images have been visually classified to determine whether each pair is matching or not. The aerial image classification step into urban/rural using CNN achieves an accuracy of 91\%. But let us recall that this classification is only an intermediate step to ease the overall change detection process.
The latter was able to identify changes with an overall accuracy of 73\%. The preclassification step has proven to be effective as if bypassed the accuracy drops to 60\%. We expect this accuracy would increase if the dataset was expanded, especially if it contains a significant amount of pairs with very slight changes between the aerial and the top-down view image labeled as unmatching.

Interesting extensions to this work would be exploring having several orientations of each top-down view image and ranking them by closest distance and picking the image ranked the highest to avoid detecting a false change due to different orientations. Road signs, lamp posts, and cars are a great cause of false positives for change detection. Detecting these objects and eliminating them from the scene would definitely affect the outcome. Finally, photometric corrections should be applied in the top-down view construction step to minimize the distortions that occur to objects far off in the scene. Indeed such distortions cause narrow places (especially in cities) lead to difficult comparison between top-down view and aerial images, as illustrated in the last line of Fig.~\ref{fig:sequenceofimages}.

%% file: PEP-ALP.tex
\label{sec:match}
Terrestrial pictures are a rich source of information for the study of landscape variations. Specifically, in studies looking at temporal evolution~\cite{Web10}, they provide unmatched spatial and temporal resolution. However, unlike standard remote sensing data, sufficiently accurate orientation is often missing, which is required to relate pixels in images with world coordinates. This limited orientation accuracy hinders their use for climate change or territorial development studies in practice.

In a database of webshared pictures, the localization of pictures is of different provenance and quality. Moving from the most to the least accurately localized source, images roughly fall into four categories:
\begin{itemize}
    \item photos taken with a GPS compass-enabled camera, having an accurate location and orientation stored in their metadata. The accuracy of the GPS position varies according to the number of satellites available and the perturbation of the signal generated by surrounding objects (such as buildings or other metallic objects). GPS-located pictures started to be massively available with the massive spread of smartphones and new generation reflex cameras.
    \item the location of the picture is entered manually by the user, through some websharing platforms that provide a web interface. The accuracy of the image location depends on the ability of the user to indicate the proper location, the resolution of the manual mapping interface, etc.
    \item the user has added some tags to the picture. Those tags may contain place names that help to roughly locate the picture.
    \item no geographic coordinates or any hint that points at a specific location are provided at all. Only the picture content can be used to compute the location.
\end{itemize}
In what follows we present a pose and orientation framework that can be deployed with pictures issued from photo-sharing web platforms. The original framework was proposed in~\cite{Pro14b}. The process has two distinct steps: in the first, we find the orientation for a set of precisely geolocated pictures in an automatic way. This is achieved by matching skylines extracted both in the picture and in a synthetic landscape rendered from a high resolution DEM. In the second step, we find both location and orientation of a second database of poorly geolocated pictures by finding appearance correspondences between the pictures and the high resolution DEM, this time augmented with the precisely oriented pictures issued from the first step.

Our goal is to meet the requirements of augmented reality applications, which do not need photogrammetric accuracy, but rather a visually good alignment of the image with a 3D model. To this end, we exploit both the geolocalisation information recorded by the camera (or provided by the user) and the 3D models stored in a GIS database.  Our proposed system helps addressing two problems of current natural landscape augmentation systems: on the one hand, the orientation provided by general public sensors (e.g. smartphones) does not reach the required accuracy; on the other hand, the matching of an image with a 3D synthetic model without geospatial constraints is difficult because state-of-the-art keypoint matching algorithms cannot be applied. This is due to the facts that i) the texture of real and synthetic images is too dissimilar (especially in the steep slopes areas) and ii) the geographical search space is too large. Hence, the solution we describe below uses this two-step logic, to benefit from the prior geolocalisation in order to geospatially constrain the matching with the 3D model.

\subsubsection{Method} 
\begin{figure}[ht]
\includegraphics[width=\linewidth]{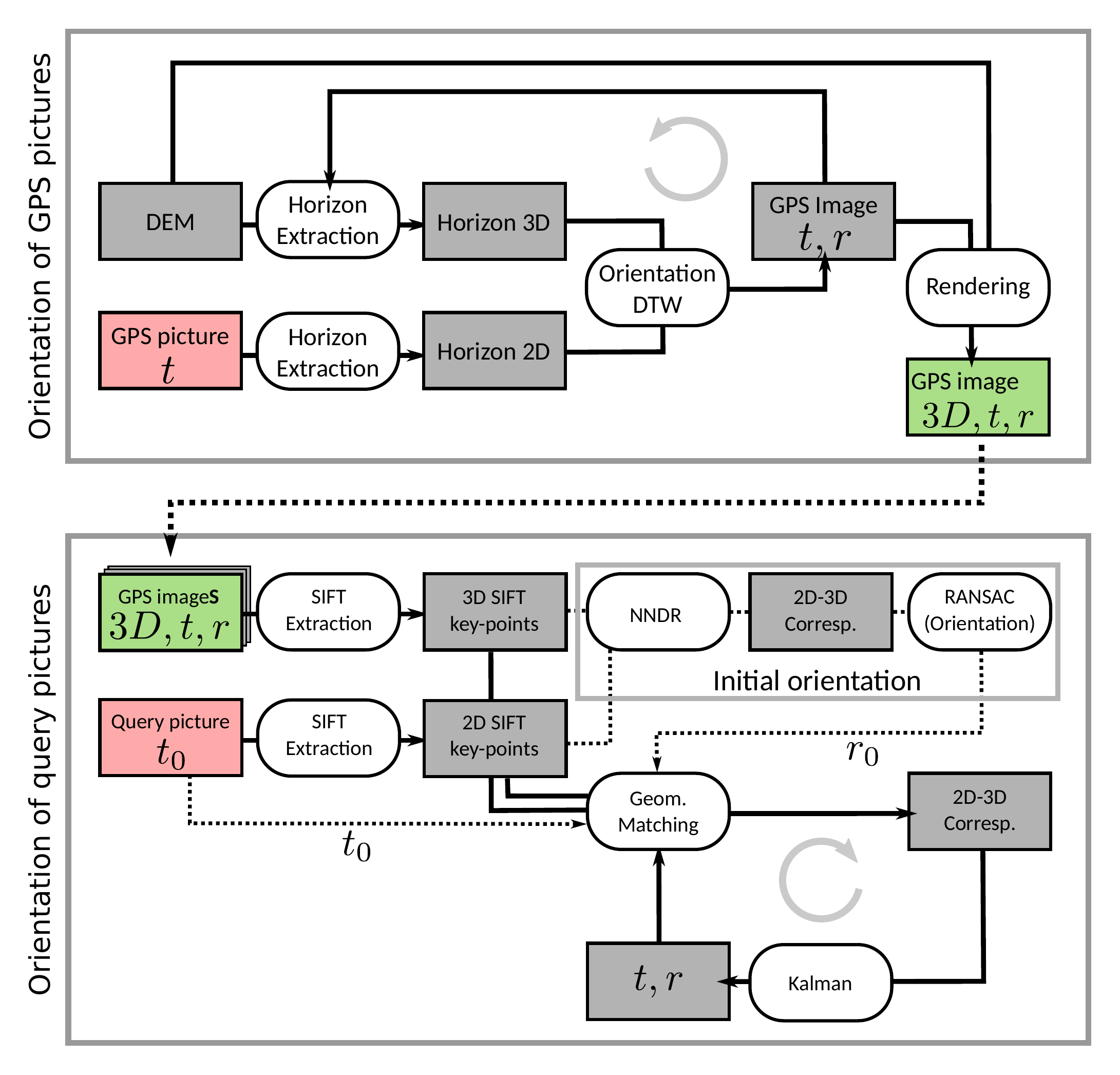}
\caption{Flowchart of the proposed method: $t$ is the vector containing the X,Y, Z coordinates describing the camera position; $r$ is the vector containing the camera angles: heading, tilt, and roll. The top schematic describes the orientation of the GPS pictures for which $t$ is known. The bottom schematic describes the orientation of the other pictures, for which prior knowledge of an approximate orientation $t_0$ and $r_0$ is assumed. 
 \label{fig:meth}}
\end{figure}

\begin{figure}[ht]
\includegraphics[width=\linewidth]{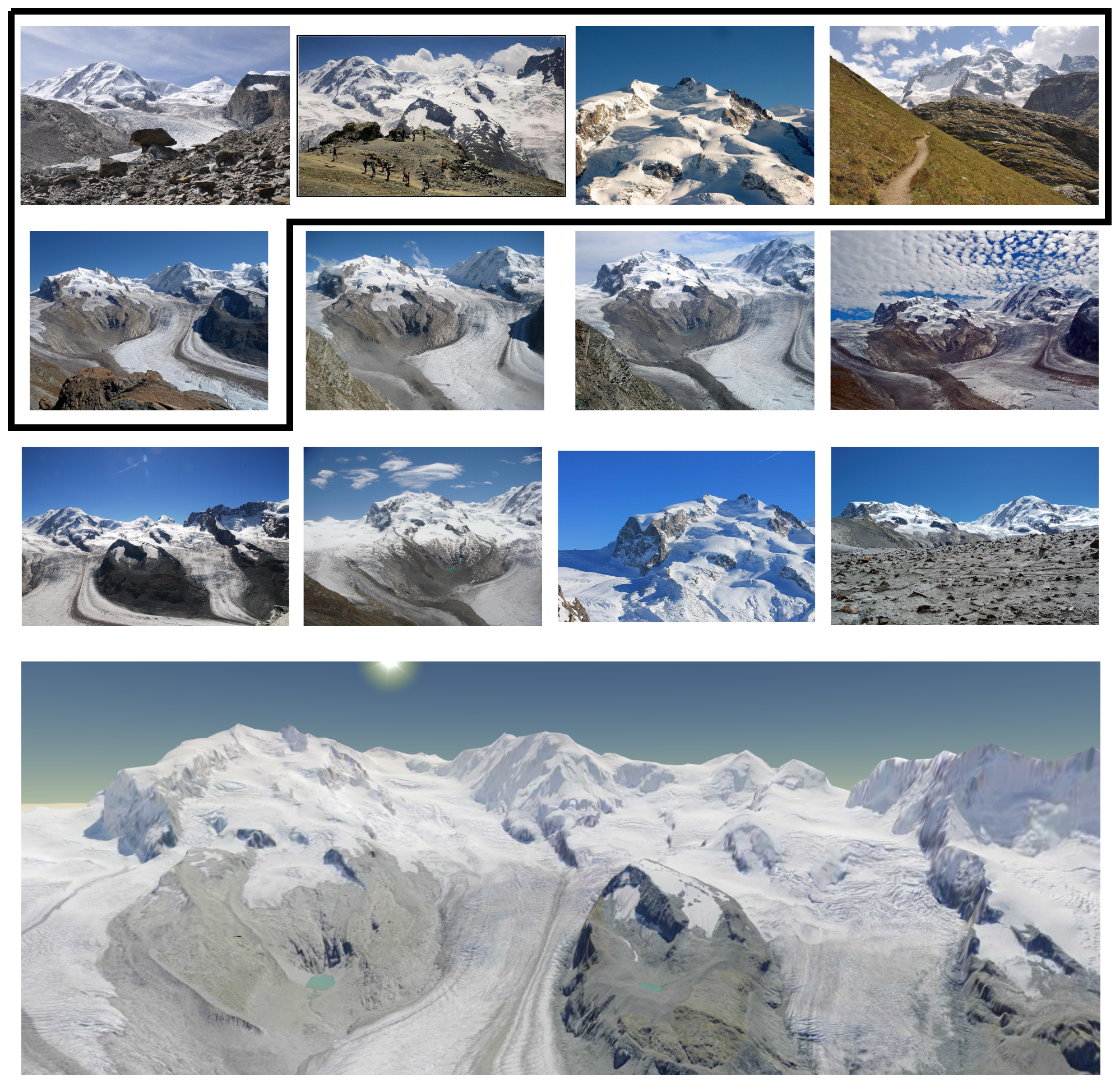}
\caption{Query pictures. The pictures inside the frame are examples of GPS-localized images. Bottom: snapshot of the reference landscape mode (DEM textured with an orthoimage). \label{fig:data}}
\end{figure}

In the following, we present the proposed pipeline, which is summarized in Fig.~\ref{fig:meth}.
The input pictures considered are representative of a webshared collection: they have various orientation information (GPS, user-provided locations or place name) and are generally clustered around easily accessible locations from which a point of interest is visible. These pictures are not yet oriented and we will call them query pictures. In the context of landscape images, a reference 3D model can be generated from a Digital Elevation Model textured with an orthoimage. Examples of the pictures and of the DEM are shown in Fig.~\ref{fig:data}.

\paragraph{Orient GPS-located pictures} the first step  (Fig.~\ref{fig:meth}, top half) aims at orienting pictures that already have a GPS position. In the context of our accuracy requirements (alignment of the image with the landscape model), we assume that the location measured by the GPS is exact. If the Field Of View (FOV) is estimated from the focal length recorded in the image metadata, only the image angles (heading, tilt and roll) must be recovered. 

We compared two methods to estimates these angles. The first estimates the orientation from the horizon silhouette. Indeed, a 360$^\circ$ synthetic panorama can be generated for a given location and the 3D silhouette can then be extracted and matched with the 2D silhouette detected in the picture to orient the camera. We developed our own method for skyline alignment, but a large range of studies exist in this field~\cite{Baa12, Bab11}. In our method, we use Dynamic Time Warping (DTW~\cite{Ber94}) to extract correspondences of the picture-extracted skyline and the reference skyline.

The second method correlates patches of the synthetic panorama image with patches extracted from the query picture. Both the query picture and the reference panorama patches are described with a HOG descriptor at several scales and matched by a correlation-based distance. 

Once GPS-located query images have been oriented, they are augmented with three bands recording the X,Y and Z coordinates of the pixels. These pictures can then be used to texture the DEM and thus improve the landscape model and serve as reference images for the next step.

\paragraph{Estimate location and orientation of poorly referenced pictures} 
the second step (Fig.~\ref{fig:meth}, bottom half) matches the remaining query images, which have a less accurate geolocalisation (e.g. the one indicated by the user), to the textured landscape model obtained at Step 1. We developed an algorithm for the Pose Estimation with Pose and Landscape Model Priors (PEP-ALP)~\cite{Pro14b}. This algorithm is based on a Kalman filter able to fuse the apriori orientation parameters (pose prior) and the 2D-3D correspondences provided by the matching of the query pictures with the reference landscape model (landscape model prior). PEP-ALP is based on the following steps: first, we use the collinearity equations~\cite{kraus2007} and variance propagation to draw in the query pictures a series of confidence ellipses. These ellipses are drawn around 3D keypoints extracted from both the reference images and the landscape model and delineate the region in which the 2D correspondence of a 3D keypoint is expected to be found. Then, keypoints from the query picture are matched to those in the landscape model. Using this geometric constraint (i.e. the confidence ellipses) decreases the number of potential correspondences and discards false positives. This is because keypoints not falling within an ellipse are removed. Finally, the Kalman filter is an iterative process, so the detected 2D-3D correspondences update the orientation estimation and the orientation covariance. At each step, the size of the ellipses is reduced together with the matching threshold, which results in an improved  estimate of the orientation.

Note that if the query picture only has an approximate location (such as one indicated by the user at small zoom or derived from a geotag), a prior orientation must be computed before applying PEP-ALP. To this end, the query picture can be matched with 3D keypoints extracted from the reference pictures located in the spatial neighborhood. If a small overlap exists with the reference images, a state-of-the-art RANSAC algorithm can extract a set of correspondences, providing the initial orientation. This orientation is then used as prior for PEP-ALP.

\subsubsection{Results and discussion}

In the following, we present two applications of the proposed workflow to picture collections of the Swiss Alps.

\paragraph{Orientation of GPS images with horizon and textured model matching}  the orientation of the GPS images was tested on a set of images in the Swiss Alps acquired by the authors. In a set of 100 images, both the horizon matching and the textured model matching were able to recover 80\% of the azimuth within 5$^{\circ}$. The failures can be explained by several reasons: on the horizon matching side, it cannot be applied if the weather is not clear or when  the mountain peaks are hidden (typically by foreground object closer to the camera), but this morphologic feature is relatively unaffected by lightning and seasonal variations. On the the synthetic panorama side, the matching is less dependent on the horizon visibility, but is more impacted by seasonal variations that may change the land cover and harm matching based on keypoints.

\begin{figure}[ht]
\includegraphics[width=\linewidth]{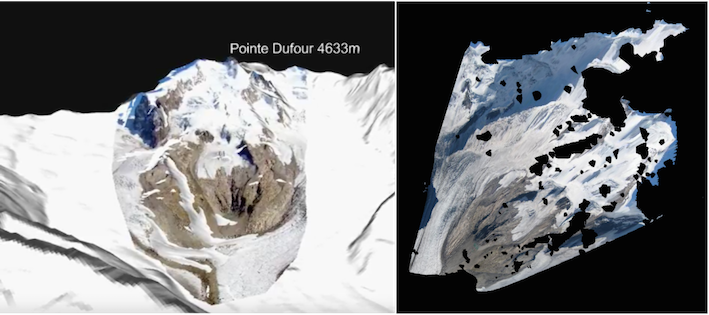}
\caption{Left: Oriented image textured on the DEM. Right: Oriented image orthorectified. \label{fig:pepalp}}
\end{figure}

\paragraph{Orientation of a real collection of webshared images} we tested the approach with a set of pictures downloaded from Panoramio in a famous Swiss alpine landscape (Zermatt). In total, the set of images in the area of interest is composed of 198 images, among which 10 have a GPS location and 118 have the focal stored in the metadata. Some examples of the pictures considered, as well as the landscape model, are shown in Fig.~\ref{fig:data}. The initial reference landscape model consists of a DEM at 25m resolution textured with an orthoimage at 0.5m resolution.
The orientation quality was assessed in two ways. First, we measured the distances between remarkable points of the orthoimage and query pictures in the geographical space. Second, the oriented pictures were projected on the DEM to render a virtual model. This model allowed us to visually assess the alignment of the pictures with the DEM. An illustrative video can be found on \url{https://youtu.be/87dHVDdlPSs}, see Fig.~\ref{fig:pepalp}.

The orientation of GPS-enabled pictures performs remarkably well if the GPS position is correct and the skyline is not occluded. Once the pose and the 3D coordinates have been estimated for all the GPS located ones, we can use them as references for the remaining query pictures. A pose is computed for the query pictures, for which RANSAC can found initial matches for (i.e. 100 images over the 188 query pictures). However, some poses are clearly incorrect (10 images): these incorrect poses are associated to images, for which RANSAC returns an incorrect initial pose (based only on false positives).
By simulating priors at various distances of the ground truth position of a picture, we showed that the success of PEP-ALP in orienting query pictures depends on the quality of the initial orientation guess. If such guess is not within 1km from the correct one, the pose and orientation estimation is seldom successful.

In future work we will fuse more intimately the three matching techniques (horizon, reference GPS images and textured landscape model). In the current implementation, the textured landscape model matching operates only in the orientation of the GPS pictures. However, the encouraging results obtained so far suggest to involve it also in the orientation of the remaining query pictures.

%% file: registree.tex
The \textit{RegisTree} project\footnote{\url{http://vision.caltech.edu/registree/}, further project members are Steve Branson, David Hall, Pietro Perona (all Caltech), and Konrad Schindler (ETH Zurich)} aims at developing a fully automated street tree monitoring pipeline that can process thousands of trees within a few hours using publicly available aerial and street-level images. It is designed as an automated, image-based system that detects trees, recognizes their species, and measures further parameters like the trunk diameter at breast height (DBH). The system is centered around state-of-the-art supervised deep convolutional neural networks~\cite{girshickICCV15fastrcnn,shaoqing15fasterRcnn,simonyan2015,szegedy2015}, where detections from multiple views are combined with a (piece-wise learned) discriminative probabilistic framework~\cite{wegner2016}. 

Trees are of uttermost importance for quality of life in cities\cite{mcpherson2016}. A healthy canopy cools cities, decreases energy demand, prevents soil erosion, slows rain-water runoff, and is key to clean and ample water supply. However, the amount of trees, their exact location, species, age, health and further parameters are often unknown because no up-to-date database exists. Today, the position, species and further attributes of trees are usually acquired manually in the field, which is labor-intensive and costly. Since the cost of employing trained arborists is prohibitive, manually acquired inventories are also not always as granular and reliable as hoped.

Automated tree mapping and species recognition has been an important research topic for the last decades (refer to~\cite{larsen2011,kaartinen2012} for a detailed comparison of methods). 
Most works follow the standard classification pipeline: extract a (relatively small) set of texture and shape features from images and/or LiDAR data, and train a classifier (e.g., Linear Discriminant Analysis, Support Vector Machines) to distinguish between a low number of species (3 in~\cite{leckie2005,heikkinen2011,korpela2011}, 4 in~\cite{heinzel2012}, 7 in~\cite{waser2011,pu2012}). Moreover, most approaches rely on data like full-waveform LiDAR, hyperspectral imagery, or very high-resolution aerial images that require dedicated, expensive flight campaigns, which limits their applicability in practise. 

\begin{figure*}[ht!]
\centering
\includegraphics[width=1.0\textwidth]{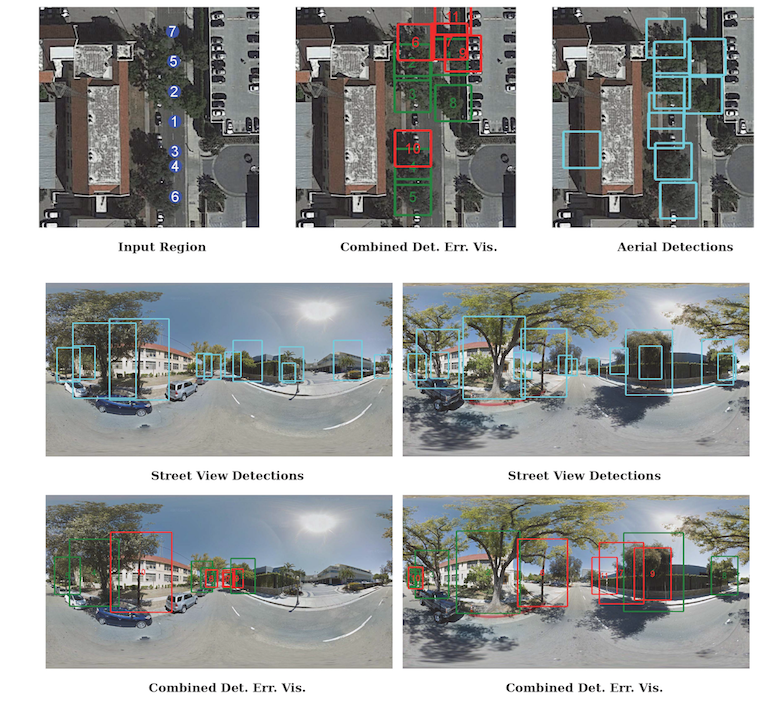}
\caption{Example for tree detection results (based on supplementary material of~\cite{wegner2016}): (top left) input aerial image with camera positions of the street-level panoramas (3 (left) and 5 (right) are shown below), (top center) aerial image overlaid with true positives (green) and false positives (red), (top right) tree detections based only on the aerial image, (middle left) tree detections based only on panorama 3, (middle right) tree detections based only on panorama 5, (bottom left) true positives (green) and false positives (red) overlaid to panorama 3 and (bottom right) panorama 5.}
\label{fig:tree_det_results}
\end{figure*}

\subsubsection{Method}

The idea here is to avoid application-specific data, and solely rely on publicly available, standard RGB images. Missing spectral channels (e.g., near-infrared), that usually contribute significant evidence to tree detection and species recognition, are compensated by the sheer amount of training data in combination with powerful, end-to-end trained deep learning models. Publicly available imagery has already found its use in a great number of applications and circumstances (e.g.,~\cite{Hay08,agarwal2009,matzen2014}). However, the process of cataloguing and classifying publicly visible objects (trees, but also street signs, solar panels, mail boxes, etc.) is still carried out by hand, often by in-person inspection or using expensive sensors as surveying-grade (LiDAR-based) mobile mapping platforms. Due to the cost, time, and organizational headache it involves, such information is rarely collected and analyzed. 
The hope is that harvesting such information automatically from publicly available street-level and aerial imagery will provide inexpensive, reliable and ready-to-use information to administrators, scientists and the general public, leading to important improvement in the quality and timeliness of public resource management. The system consists of a tree detection component and a species classifier (see details in~\cite{wegner2016}).

\paragraph{Tree detection}

Faster R-CNN~\cite{shaoqing15fasterRcnn} is applied to detect trees in all available street-level and aerial images, that are downloaded densely for each specified area. Note that establishing correspondence across aerial and street-level images generally is a wide-baseline matching problem. Because of the drastic change of view-point, direct dense pixel-wise matching and even sparse matching of key points is largely unsolved. Although promising approaches for ground-to-aerial image matching (for buildings) have been proposed recently (e.g., \cite{shan2014,Lin2015}), matching trees that are non-rigid mostly, and change appearance depending on the season, is literally impossible. We thus avoid point-wise correspondence search altogether, and resort to late fusion of detection results. Individual tree detection is done separately per image and all individual detections are combined via soft probabilistic fusion in geographic coordinates\footnote{While transfer from image bounding boxes to geographic coordinates is trivial for orthophotos, we assume locally flat terrain (and known camera height) for single street-level panoramas.}. 

In more detail, the following is done:
i) Run the Faster R-CNN detector with a liberal detection threshold to compute an over-complete set of region proposals per view;
ii) Project detections of all views to geographic coordinates;
iii) Collect all detections in geographic coordinates in a multi-view region proposal set by taking the union of all individual view proposals;
iv) Project each detection region back into each view.
v) Evaluate all detection scores in each view of the combined multi-view proposal set;
vi) Compute a combined detection score over all views, apply a detection threshold, and suppress overlapping regions to obtain geographic detections using a probabilistic conditional random field (CRF) approach.
This workflow is robust to initially missing detections from single views because it collects all individual detections in geographic space and projects these back to all views, i.e. scores are evaluated also in views where nothing was detected in the first place. The CRF framework also allows adding prior information to pure detection scores. 

A spatial context potential encodes prior knowledge about the usually very regular spacing of neighboring trees on the roadside, that have been artificially planted by city administration to the largest part. However, the prior could cope with any kind of distribution of neighboring objects because it is learned from the training data set. 
Another prior based on map data models the distribution of distances between trees and the road boundary. It helps suppressing false positives by encoding that trees rarely grow in the middle of a road but are usually planted alongside at a certain distance. Tree-to-road-edge distances are computed per pixel using downloaded Google maps.

\subsubsection{Tree species recognition}

Given geographic coordinates of tree objects, the goal is to predict each trees species. For this purpose we again use state-of-the-art CNNs, in this case the GoogLeNet CNN model~\cite{szegedy2015}. One aerial image and three cropped versions at increasing zoom level of the closest street-view panorama are collected per tree. Four separate GoogLeNet models are trained, features of the four models are stacked into a single feature vector per object, and a linear SVM is finally trained.  

\subsubsection{Results} First results of the \textit{RegisTree} project~\cite{wegner2016} show that the system is able to detect $>70\%$ of all trees in the city of Pasadena (CA), and to correctly classify $\approx80\%$ of those trees into the 18 species that make up most of the town’s population. Tree detection results for an example scene are shown in Fig.~\ref{fig:tree_det_results}. Red boxes highlight typical cases for false positives, which are usually due to either (10) wooden utility poles and (rarely) street signs (11), (9) double detections of the same tree at two different positions (1, correct) and (9, wrong). The latter case is usually due to inaccurate projection of detections in images to geographic coordinates. This can be caused by inaccurate GPS and heading information that comes with the panorama images, a violation of the locally flat terrain assumption, or simply wrong bounding box position and size in the image, which seems to be the reason for error (9) here.  
A first larger-scale feasibility study for all of the Los Angeles bay area, shows that the system is able to detect and classify more than a million trees from ca.~500 different species (data verification pending)\footnote{For an interactive demo, check the project webpage \url{http://vision.caltech.edu/registree/explore-demos.html}.}.

One insight of this project is that state-of-the-art deep learning methods in combination with very large-scale, standard RGB imagery, and a massive amount of publicly available ground truth can lead to surprisingly good results. The massive amount of training data over-compensates the partially poor quality of the reference data. Soft probabilistic fusion of detector outputs of multiple viewpoints (under simplifying but often sufficiently fulfilled assumptions like locally flat terrain) can help avoiding direct per-pixel or per-key point matching. Such a detector-based, soft probabilistic fusion approach would also allow the introduction of different modalities like LiDAR or SAR, for example, as long as correspondence can be established via geographic coordinates.

%% file: conclusion.tex
In this paper, we have presented a thorough, comprehensive review and discussion of works that tackle the challenging but very promising task of combining images from satellite and aerial to street-level sensors. It turns out that there is not one gold-standard or a single, general framework that can generically be applied to any kind of task or dataset. Rather, different applications and data scenarios call for task-specific techniques. 
A common hurdle is the very wide baseline between overhead imagery and images acquired on the ground. This drastic viewpoint change makes establishing direct, dense correspondence on per-pixel level largely impossible and one has to resort to alternative strategies. Thus, multiple viewpoints are rarely combined at an early image-stage, but rather at a later stage on per-object basis, through intermediate representations, or relying on late fusion of separate detector outcomes. This implies that many state-of-the-art methods combine knowledge of the sensor geometries with machine learning to align detections across viewpoints.

The ever-increasing amount of viewpoints is likely to ease the wide-baseline matching problem in the near future. For example, oblique images are today often captured by aerial campaigns, while 3D information can be reconstructed from stereoscopic images acquired with EO satellite sensors. These various oblique images can serve to bridge the gap between nadir imagery and street-level data. Great promise in this regard are also UAVs, which allow photo acquisition from basically arbitrary viewpoints. This heterogeneous but rich collection of multi-view data in combination with powerful machine learning techniques can be viewed as a unified approach to model our environment.